\documentclass[letterpaper, 10 pt, conference]{ieeeconf}  

\usepackage{amsmath}
\usepackage{graphics} 
\usepackage{epsfig} 
\usepackage{mathptmx} 
\usepackage{times} 
\usepackage{amssymb}  
\usepackage{booktabs}
\usepackage{multirow}
\usepackage{multicol}
\usepackage{tabularx}
\usepackage{tablefootnote}
\usepackage{xcolor}
\usepackage{makecell}
\usepackage{url}
\usepackage[table]{xcolor}

\DeclareMathAlphabet{\mathcal}{OMS}{cmsy}{m}{n}

\newcolumntype{Y}{>{\centering\arraybackslash}X}

\IEEEoverridecommandlockouts                              

\title{\LARGE \bf
LIFT: \underline{L}ocal Metric Al\underline{i}gnment and
\underline{F}oundation-guided Propaga\underline{t}ion \\for
Few-shot Depth Completion
}

\author{Xiang Li and Xiaojin Gong$^{*}$%
\thanks{Xiang Li and Xiaojin Gong are with the College of
Information Science and Electronic Engineering,
Zhejiang University, Hangzhou, Zhejiang 310027, China
(e-mail: \texttt{lx021101@gmail.com};
\texttt{gongxj@zju.edu.cn}).}%
\thanks{$^{*}$Corresponding author.}%
\thanks{This work has been submitted to IEEE for possible
publication. Copyright may be transferred without notice,
after which this version may no longer be accessible.}%
}

\begin{document}

\maketitle
\thispagestyle{empty}
\pagestyle{empty}

\begin{abstract}
Few-shot depth completion aims to recover dense metric depth from sparse measurements using only a small number of  target-domain annotations, offering a favorable balance between annotation cost and model adaptation. We propose LIFT, a framework that harnesses both depth and feature priors from a frozen pretrained monocular depth estimation (MDE) foundation model for few-shot depth completion. Specifically, LIFT first estimates local scale and shift parameters on a coarse grid and interpolates them into smooth, dense parameter fields, providing a spatially adaptive metric depth initialization. It then progressively refines this initialization through multi-scale depth propagation guided by foundation features. Finally, a lightweight residual decoder combines multi-level foundation features with propagation correction to refine the metric depth. Extensive experiments on NYUv2 and KITTI-DC demonstrate that our proposed method achieves state-of-the-art few-shot performance.
The source code will soon be released.


\end{abstract}

\section{INTRODUCTION}
Depth completion aims to recover dense metric depth from sparse depth measurements with the aid of an RGB image. Compared with other 3D vision paradigms, such as Monocular Depth Estimation (MDE)~\cite{yao2025monocular} and stereo vision~\cite{schops2017multi}, it can provide more reliable 3D geometry by leveraging direct depth observations, particularly in visually challenging scenarios. This advantage has made depth completion a fundamental task in autonomous driving~\cite{song2021self} and robotics~\cite{ma2019sparse}, where accurate metric geometry remains essential for downstream perception and decision-making.

With the advent of foundation models, particularly large-scale MDE foundation models~\cite{birkl2023midas,yin2023metric3d,yang2024depth1,yang2024depth2,ke2024marigold,gui2025depthfm}, depth completion has increasingly shifted from designing task-specific models from scratch toward leveraging powerful pretrained priors. Existing foundation model-driven approaches can be broadly categorized into training-based and test-time adaptation methods. Training-based methods~\cite{lin2025prompting,gregorek2026need,wang2026depth} typically incorporate pretrained MDE models into dedicated architectures and train them on large-scale synthetic datasets, such as Hypersim~\cite{roberts2021hypersim} and Virtual KITTI~\cite{gaidon2016virtual, cabon2020virtual}, before transferring to real-world benchmarks. While effective, they require substantial training data and computational resources. In contrast, test-time adaptation methods~\cite{viola2025marigold,gregorek2025steeredmarigold,hyoseok2025zero,jeong2025test,seo2026efficient,ke2026depth} adapt pretrained MDE models to sparse depth observations during inference without task-specific training. However, their reliance on iterative optimization for each test sample introduces considerable inference overhead, limiting their practicality for real-time robotic applications.

These limitations motivate a more efficient few-shot paradigm that leverages pretrained MDE priors for label-efficient depth completion, as explored by UniDC~\cite{park2024simple} and OASIS-DC~\cite{cho2026oasis}. Following this direction, we investigate how to better exploit pretrained foundation priors under limited supervision for accurate depth completion. Regardless of whether these priors are leveraged through few-shot adaptation or training-based approaches, two key challenges remain: \emph{how to align dense relative depth with sparse metric measurements, and how to fully exploit the rich semantic and geometric priors encoded in monocular foundation models}.
 
For the first challenge, various alignment strategies have been developed based on global alignment~\cite{viola2025marigold,gregorek2026need,zhou2026any}, Poisson-based optimization~\cite{yu2026large,cho2026oasis}, and pixel-wise local fitting~\cite{wang2026depth}. However, global alignment cannot capture spatially varying discrepancies, while Poisson-based optimization does not explicitly model local scale and shift variations. Pixel-wise local fitting, in contrast, may suffer from unstable parameter estimation under sparse observations. To address these limitations, we propose a grid-based local metric alignment strategy, which estimates local transformation parameters on a coarse regular grid using nearby sparse measurements and interpolates them to the full resolution. This design captures spatially varying metric discrepancies while providing stable and efficient alignment.

For the second challenge, the aligned depth prior alone may not fully exploit the rich information encoded in pretrained foundation models. To this end, we introduce multi-scale foundation feature-guided depth propagation, which leverages multi-level features from a frozen MDE encoder to guide the propagation of reliable depth information within semantically and geometrically consistent regions. A lightweight residual decoder further refines the propagated depth to correct remaining fine-grained discrepancies. Together, these components effectively exploit the complementary depth and feature priors of pretrained foundation models for accurate metric depth recovery under few-shot supervision.

Our contributions are summarized as follows:
\begin{itemize}
		\item We introduce LIFT, a foundation-prior-based framework that achieves state-of-the-art performance under few-shot supervision on both indoor and outdoor datasets.

    \item We propose a simple yet effective local alignment strategy that captures spatially varying metric discrepancies and provides stable alignment under sparse metric observations.
    
    \item We develop multi-scale foundation feature-guided depth propagation with lightweight residual refinement to effectively exploit semantic and geometric priors for dense metric depth recovery.
\end{itemize}


\section{Related Work}
\subsection{Monocular Depth Foundation Models}
Recent vision foundation models have significantly advanced MDE through large-scale pretraining and powerful visual representations. Existing monocular depth foundation models can be broadly categorized into discriminative and generative paradigms. Discriminative models, such as the Depth Anything series~\cite{yang2024depth1,yang2024depth2,lin2025depth}, perform efficient dense regression, while generative models, such as Marigold~\cite{ke2024marigold} and DepthFM~\cite{gui2025depthfm}, leverage diffusion priors to recover finer geometric details. Despite their strong generalization, most foundation models predict scale-ambiguous relative depth, motivating metric MDE methods such as ZoeDepth~\cite{bhat2023zoedepth} and Metric3D~\cite{yin2023metric3d,hu2024metric3d}. However, recovering absolute scale from a single image remains challenging, often resulting in scale errors and geometric distortions in metric MDE.

\subsection{Foundation-driven Depth Completion}
Foundation-driven depth completion leverages pretrained MDE priors together with sparse sensor measurements for accurate depth reconstruction. Existing methods can be categorized into training-time and test-time paradigms. Training-time methods exploit pretrained MDE models as geometric priors and learn to align or fuse them with sparse measurements. Representative approaches, including PromptDA~\cite{lin2025prompting}, Any2Full~\cite{zhou2026any}, Marigold-SSD~\cite{gregorek2026need}, PriorDA~\cite{wang2026depth}, and LDCM~\cite{yu2026large}, rely on substantial training with synthetic or real-world data. Several of these methods demonstrate zero-shot generalization to unseen real-world domains, although domain gaps remain a challenge. In contrast, test-time methods, such as Marigold-DC~\cite{viola2025marigold}, SteeredMarigold~\cite{gregorek2025steeredmarigold}, Zero-DC~\cite{hyoseok2025zero}, TestPromptDC~\cite{jeong2025test}, LRDA~\cite{seo2026efficient}, and CAPA~\cite{ke2026depth}, adapt pretrained models during inference using scene-specific sparse observations, avoiding extensive training at the expense of additional inference overhead. To balance training cost and inference efficiency, few-shot methods, including UniDC~\cite{park2024simple}, OASIS-DC~\cite{cho2026oasis}, and our approach, adapt foundation-model priors with only a small number of target-domain RGB-depth pairs.

\subsection{Metric-Relative Depth Alignment}
Most MDE foundation models predict scale-ambiguous relative depth, making metric-relative depth alignment a fundamental step in foundation-driven depth completion. Existing methods either estimate a single global affine transformation from sparse observations, as in Marigold-DC~\cite{viola2025marigold}, Marigold-SSD~\cite{gregorek2026need}, and Any2Full~\cite{zhou2026any}, or enforce local geometric consistency through Poisson-based optimization, as in LDCM~\cite{yu2026large} and OASIS-DC~\cite{cho2026oasis}. While global alignment cannot capture spatially varying calibration errors, Poisson-based methods require additional optimization and primarily enforce gradient consistency. To enable spatially adaptive alignment, PriorDA~\cite{wang2026depth} performs pixel-wise KNN-based local least-squares fitting, but independently estimated parameters can fluctuate across neighboring pixels due to varying sparse neighborhoods. In contrast, we formulate alignment as a continuous parameter field, estimating local transformation parameters on a coarse spatial grid and densely reconstructing them via bilinear interpolation, thereby preserving spatial adaptivity while improving consistency.

\subsection{Spatial Propagation}
Spatial propagation has been widely used in depth completion to refine dense predictions through affinity-guided neighborhood aggregation. Existing methods mainly differ in how propagation affinities are obtained. CSPN~\cite{cheng2018depth}, CSPN++~\cite{cheng2020cspn++}, NLSPN~\cite{park2020non}, and DySPN~\cite{lin2023dyspn} learn propagation affinities using dedicated networks, with subsequent works further introducing adaptive neighborhoods or dynamic propagation. UniDC~\cite{park2024simple} and OASIS-DC~\cite{cho2026oasis} further exploit hyperbolic representations to construct geometry-aware propagation affinities. Building on this line of work, our method computes local affinities from projected features of a frozen MDE encoder and uses them for depth aggregation within a coarse-to-fine refinement pipeline.



\section{Method}
\begin{figure*}[!t]
\centering
\includegraphics[width=1.0\linewidth]{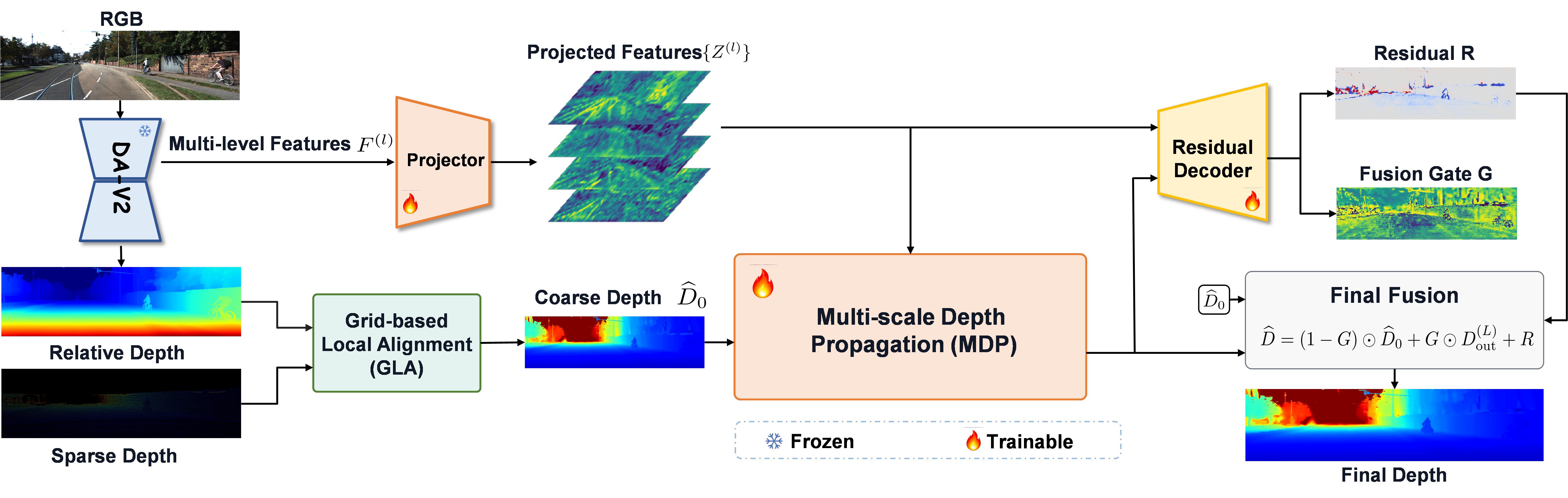}
\caption{Overview of the proposed LIFT framework. Given an RGB image and sparse depth, the frozen MDE foundation model provides relative depth and multi-level features. The relative depth is first converted into a coarse metric depth by the Grid-based Local Alignment (GLA) module, then refined by the Multi-scale Depth Propagation (MDP) module guided by foundation features, and finally enhanced by a lightweight residual decoder. The final metric depth is obtained by combining the coarse metric depth, propagated depth, and predicted residual.}
\label{fig:framework}
\end{figure*}

Given an RGB image $I\in\mathbb{R}^{H\times W\times 3}$ and a sparse depth map $D_s\in\mathbb{R}^{H\times W}$, our goal is to recover dense metric depth $\widehat{D}\in\mathbb{R}^{H\times W}$ using only a few annotated training frames while fully leveraging the priors of a pretrained monocular depth estimation (MDE) foundation model. We adopt the frozen Depth Anything V2~\cite{yang2024depth2} as the MDE model, which provides both a relative depth map $D_{\mathrm{r}}$ and multi-level features. Building upon these foundation priors, we propose LIFT,  a depth completion framework composed of three components: a Grid-based Local Alignment (GLA) module for spatially adaptive metric alignment, a Multi-scale Depth Propagation (MDP) module guided by MDE features to improve semantic consistency, and a lightweight residual decoder for recovering fine-grained details. Together, these components progressively transform the relative depth prior into a dense metric depth prediction. An overview of the proposed framework is shown in Fig.~\ref{fig:framework}.

\subsection{Grid-based Local Alignment}
\begin{figure}[!tb]
\centering
\includegraphics[width=1.0\linewidth]{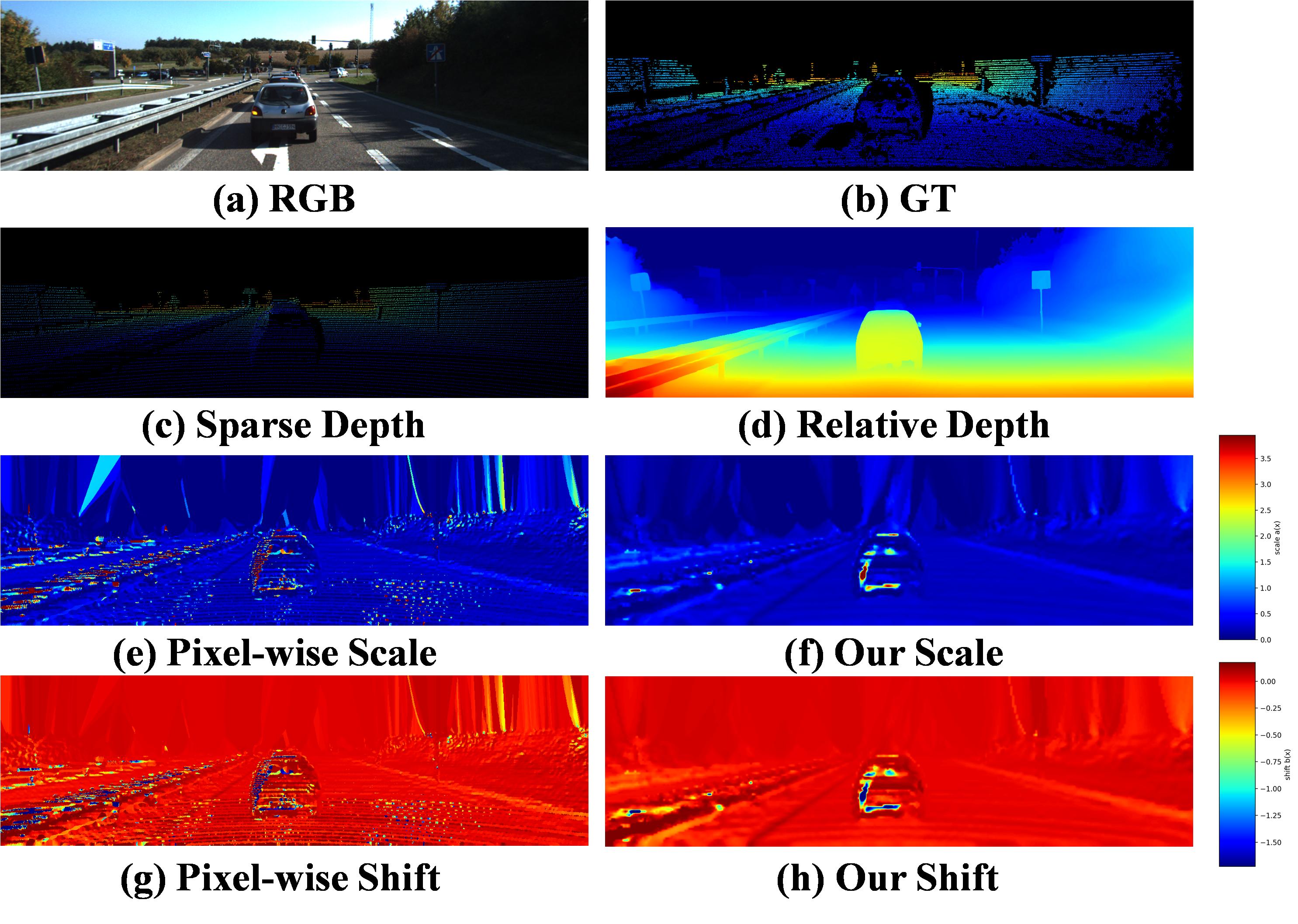}
\caption{Illustration of the scale and shift fields estimated by pixel-wise local fitting and our grid-based local alignment method.}
\label{fig2}
\end{figure}

Metric-relative depth alignment reconciles the scale-ambiguous relative depth predicted by MDEs with sparse metric observations. Considering that global fitting~\cite{viola2025marigold,gregorek2026need,zhou2026any} cannot capture spatially varying alignment errors, PriorDA~\cite{wang2026depth} adopts pixel-wise local fitting. However, independently estimating affine parameters at every pixel may lead to unstable parameter fluctuations under sparse and noisy observations, as illustrated in Fig.~\ref{fig2} (e) and (g). To better balance spatial adaptability and parameter stability, we propose a simple yet effective design, Grid-based Local Alignment (GLA). GLA performs local fitting at regularly spaced locations and bilinearly interpolates the resulting parameters to the full resolution. This design constructs a smooth parameter field from fewer local regressions, enabling spatially adaptive alignment with more stable parameter estimation, as shown in Fig.~\ref{fig2} (f) and (h).

Specifically, GLA first defines a set of query points $\Omega_q$ on a coarse spatial grid. For each query point $q=(x_q,y_q)\in\Omega_q$, we select the $k$ nearest valid sparse metric observations on the image plane, denoted by $\mathbf{N}_k(q)$. The local scale $s_q$ and shift $t_q$ are then estimated by solving the following weighted least-squares problem:
\begin{equation}
(s_q,t_q)
=
\arg\min_{s,t}
\sum_{p\in\mathbf{N}_k(q)}
\frac{
\left(
s\bar{D}_{\mathrm{r}}(p)+t-D_s(p)
\right)^2
}{
\lVert q-p\rVert_2
},
\label{eq:gla_lsq}
\end{equation}
where $D_s(p)$ denotes the sparse metric depth at location $p=(x_p,y_p)$, $\bar{D}_{\mathrm{r}}$ is the per-image min--max normalized version of the relative depth map $D_{\mathrm{r}}$.
 
The estimated $\{s_q,t_q\}$ values form coarse scale and shift fields, which are bilinearly interpolated to the original image resolution to obtain dense parameter fields $S\in\mathbb{R}^{H\times W}$ and $T\in\mathbb{R}^{H\times W}$. The aligned metric depth is then obtained by
\begin{equation}
\widehat{D}_0 = S \odot \bar{D}_{\mathrm{r}} + T,
\label{eq:gla_depth}
\end{equation}
where $\odot$ denotes element-wise multiplication.

\subsection{Multi-scale Depth Propagation}
While GLA provides spatially adaptive metric alignment, local errors and object-level inconsistencies may still remain due to sparse measurements and parameter interpolation. Meanwhile, the DINOv2-based~\cite{oquab2024dinov2} encoder of Depth Anything V2~\cite{yang2024depth2} naturally encodes rich semantic and geometric cues. Considering that higher-level features encode stronger semantic context while lower-level features preserve finer local structures, we perform depth propagation in a coarse-to-fine manner. Metric depth is progressively propagated across feature scales, enabling long-range semantic guidance while preserving local geometric continuity.

Let $F^{(l)}\in\mathbb{R}^{H'\times W'\times C_l}$ denote the feature map extracted from the encoder layer assigned to the propagation stage $l$, where $H'$, $W'$, and $C_l$ denote the height, width, and channel dimension, respectively. Each feature map is projected into a 64-dimensional embedding and bilinearly resized to the corresponding propagation resolution. We distinguish the resized features from their channel-wise \(\ell_2\)-normalized counterparts:
\begin{equation}
Z^{(l)}
=
\operatorname{Resize}_{H_l,W_l}
\left(\phi_l(F^{(l)})\right),
\qquad
\bar F^{(l)}
=
\operatorname{Norm}_2\left(Z^{(l)}\right),
\end{equation}
where \(Z^{(l)},\bar F^{(l)}\in\mathbb R^{H_l\times W_l\times64}\), and \(\phi_l\) is a trainable projector consisting of channel-wise LayerNorm, a \(1\times1\) convolution, GroupNorm, and GELU. Here, \(\operatorname{Resize}_{H_l,W_l}\) denotes bilinear interpolation to the propagation resolution, and \(\operatorname{Norm}_2\) denotes \(\ell_2\) normalization along the channel dimension. The normalized features \(\bar F^{(l)}\) are used to compute affinities, while \(Z^{(l)}\) is used by the propagation gates and residual decoder.

For a query pixel $p$ and its neighbor $q\in\mathbf{N}(p)$, we compute
the feature affinity as
\begin{equation}
w^{(l)}_{pq}
=
\frac{
\exp\!\left(
\bar{F}^{(l)}(p)^{\top}\bar{F}^{(l)}(q)
\right)
}{
\displaystyle
\sum_{q'\in\mathbf{N}(p)}
\exp\!\left(
\bar{F}^{(l)}(p)^{\top}\bar{F}^{(l)}(q')
\right)
}.
\label{eq:affinity}
\end{equation}
Here, $\mathbf{N}(p)$ is defined as a local $3\times3$ window at the $l$-th feature scale. Its effective spatial coverage varies across scales, capturing broader context at coarse levels and finer structures at fine levels. This compact neighborhood also keeps the propagation computationally efficient while preserving local geometric continuity.

To adaptively control depth propagation, we introduce a learnable spatial gate:
\begin{equation}
g^{(l)}
=
\sigma\!\left(
\operatorname{Conv}_{3\times3}\!\left(
Z^{(l)},D^{(l)}_{\mathrm{in}}-\widehat{D}^{(l)}_0
\right)
\right),
\label{eq:prop_gate}
\end{equation}
where $g^{(l)}(p)\in[0,1]$ controls the extent to which neighboring information updates the current depth estimate.
Here, $\operatorname{Conv}_{3\times3}$ denotes a $3\times3$ convolution, and $\sigma(\cdot)$ denotes the sigmoid function. $\widehat{D}^{(l)}_0$ denotes the coarse depth map obtained by downsampling $\widehat{D}_0$ to the feature resolution at scale $l$, and $D^{(l)}_{\mathrm{in}}$ is the input depth map at scale $l$, which will be defined below.
The depth propagation at feature scale $l$ is performed by
\begin{equation}
D^{(l)}_{\mathrm{out}}(p)
=
\left(1-g^{(l)}(p)\right)D^{(l)}_{\mathrm{in}}(p)
+
g^{(l)}(p)
\sum_{q\in\mathbf{N}(p)}
w^{(l)}_{pq}D^{(l)}_{\mathrm{in}}(q).
\label{eq:propagation}
\end{equation}
Here, the input depth $D^{(l)}_{\mathrm{in}}$ preserves the sparse metric
observations while filling the remaining regions with the coarse metric
depth at the coarsest level or the propagated depth from the preceding
scale:
\begin{equation}
D^{(l)}_{\mathrm{in}}
=
\begin{cases}
M^{(l)} \odot D_s^{(l)}
+
\left(1-M^{(l)}\right) \odot \widehat{D}^{(l)}_0,
& l=0,\\[2mm]
M^{(l)} \odot D_s^{(l)}
+
\left(1-M^{(l)}\right) \odot
\operatorname{Up}\!\left(D^{(l-1)}_{\mathrm{out}}\right),
& l>0.
\end{cases}
\label{eq:depth_initialization}
\end{equation}
Here, $D_s^{(l)}$ and $M^{(l)}$ denote the sparse depth map and its binary validity mask at scale $l$, respectively. $\operatorname{Up}(\cdot)$ denotes bilinear upsampling to the current scale.
 
\subsection{Residual Depth Refinement}
Although multi-scale depth propagation effectively enhances metric consistency and local structural coherence, fine-grained details may be lost and local artifacts may still remain due to the coarse-to-fine propagation process, particularly around object boundaries. To address these issues, we introduce a lightweight residual decoder that further refines the propagated depth and recovers fine-grained structures.

Specifically, the decoder takes the multi-scale features $\{Z^{(l)}\}$, the propagation correction $D^{(L)}_{\mathrm{out}}-\widehat{D}_0$, and the binary sparse depth mask $M$ as inputs. It progressively fuses these multi-scale cues in a coarse-to-fine manner and predicts a residual map $R$ and a fusion gate $G$:
\begin{equation}
R, G
=
\operatorname{Decoder}\!\left(
\{Z^{(l)}\},
D^{(L)}_{\mathrm{out}}-\widehat{D}_0,
M
\right).
\label{eq:decoder}
\end{equation}
The final depth prediction is obtained by combining the initial metric
estimate, the propagated depth at the finest scale, and the predicted
residual:
\begin{equation}
\widehat{D}
=(1-G) \odot \widehat{D}_0
+
G \odot D^{(L)}_{\mathrm{out}}
+
R,
\label{eq:final_depth}
\end{equation}
where $D^{(L)}_{\mathrm{out}}$ denotes the propagated depth at the
finest scale, whose spatial resolution matches that of the input image.
The fusion gate $G$ adaptively modulates the contribution of the
propagated depth, while $R$ accounts for the remaining fine-grained
depth corrections.

\subsection{Training Loss}
We optimize the propagation and refinement modules while keeping the MDE backbone frozen, using the following objective:
\begin{equation}
\mathcal{L}
=
\mathcal{L}_{\mathrm{data}}
+
\lambda_{\mathrm{g}}\mathcal{L}_{\mathrm{grad}}
+
\lambda_{\mathrm{r}}\mathcal{L}_{\mathrm{residual}}.
\end{equation}
Here, $\mathcal{L}_{\mathrm{data}}$ denotes the data fidelity loss, which combines $L_1$ and $L_2$, following common practice in depth completion~\cite{park2024simple, cho2026oasis, park2024depth}. The gradient matching loss $\mathcal{L}_{\mathrm{grad}}$
penalizes discrepancies between predicted and ground-truth depth gradients over adjacent valid pixel pairs, encouraging consistent local depth variations. Finally, $\mathcal{L}_{\mathrm{residual}}$ is an $L_1$ regularization term that encourages the predicted residual to remain sparse:
\begin{equation}
\mathcal{L}_{\mathrm{residual}}
=
\frac{1}{|\Omega|}
\sum_{p\in\Omega}
|R(p)|,
\end{equation}
where $\Omega$ denotes the set of all image pixels. The coefficients $\lambda_{\mathrm{g}}$ and $\lambda_{\mathrm{r}}$ control the relative contributions of the gradient and residual regularization terms, respectively.


\section{Experiments}
\begin{table*}[!t]
\centering
\small
\caption{Few-shot depth completion results on NYUv2 and KITTI-DC. The results of the first six methods are taken from~\cite{park2024simple,cho2026oasis}, while the remaining baseline results are reported from their respective original papers. RMSE and MAE are reported in meters. The best and second-best results for each dataset are highlighted in \textbf{bold} and \underline{underlined}, respectively.} 
\label{tab:fewshot_depth_completion}
\begin{tabular*}{\textwidth}{@{\extracolsep{\fill}}llcccccccc@{}}
\toprule
\multirow{2}{*}{Dataset}
& \multirow{2}{*}{Method}
& \multicolumn{2}{c}{1-shot}
& \multicolumn{2}{c}{10-shot}
& \multicolumn{2}{c}{100-shot}
& \multicolumn{2}{c}{1-Sequence Training} \\
\cmidrule(lr){3-4}
\cmidrule(lr){5-6}
\cmidrule(lr){7-8}
\cmidrule(lr){9-10}
& & RMSE$\downarrow$ & MAE$\downarrow$
  & RMSE$\downarrow$ & MAE$\downarrow$
  & RMSE$\downarrow$ & MAE$\downarrow$
  & RMSE$\downarrow$ & MAE$\downarrow$ \\
\midrule

\multirow{9}{*}{NYUv2}
& CSPN~\cite{cheng2018depth}
& 1.4827 & 1.2058 & 0.3166 & 0.1961
& 0.2854 & 0.1307 & 0.3166 & 0.1961 \\
& NLSPN~\cite{park2020non}
& 1.9358 & 1.6132 & 1.5995 & 0.8261
& 0.5501 & 0.4150 & 0.8881 & 0.6421 \\
& DySPN~\cite{lin2023dyspn}
& 1.5474 & 1.2851 & 0.4102 & 0.2817
& 0.3079 & 0.1706 & 0.2584 & 0.1320 \\
& CompletionFormer~\cite{zhang2023completionformer}
& 1.8218 & 1.5539 & 1.1583 & 1.0162
& 0.9914 & 0.8164 & 0.6779 & 0.5356 \\
& BPNet~\cite{tang2024bilateral}
& 0.3573 & 0.2077 & 0.2392 & 0.1120
& 0.1757 & 0.0793 & 0.2220 & 0.1040 \\
& DepthPrompting~\cite{park2024depth}
& 0.3583 & 0.2067 & 0.2195 & 0.1006
& 0.2101 & 0.1008 & 0.2335 & 0.1191 \\
& OASIS-DC~\cite{cho2026oasis}
& 0.2105 & 0.1105 & 0.1670 & 0.0838
& 0.1484 & 0.0706 & 0.1644 & 0.0787 \\
& UniDC~\cite{park2024simple}
& \underline{0.2099} & \underline{0.1075}
& \underline{0.1657} & \underline{0.0794}
& \underline{0.1473} & \underline{0.0669}
& \underline{0.1632} & \underline{0.0745} \\
& LIFT (Ours)
& \textbf{0.1449} & \textbf{0.0619}
& \textbf{0.1436} & \textbf{0.0615}
& \textbf{0.1394} & \textbf{0.0594}
& \textbf{0.1444} & \textbf{0.0620} \\

\midrule

\multirow{9}{*}{KITTI-DC}
& CSPN~\cite{cheng2018depth}
& 9.2748 & 3.5921 & 2.0222 & 0.7825
& 1.4510 & 0.5184 & 2.6289 & 0.8355 \\
& NLSPN~\cite{park2020non}
& 7.2899 & 4.7422 & 4.0070 & 2.2588
& 2.4979 & 1.1710 & 4.0290 & 1.7881 \\
& DySPN~\cite{lin2023dyspn}
& 2.6350 & 0.8870 & 2.2701 & 0.9150
& 1.8777 & 0.6188 & 2.8530 & 0.7980 \\
& CompletionFormer~\cite{zhang2023completionformer}
& 4.7212 & 2.3789 & 3.1601 & 1.4740
& 2.6122 & 1.3299 & 4.5588 & 1.9603 \\
& BPNet~\cite{tang2024bilateral}
& 5.4000 & 1.0740 & 1.8799 & 0.5559
& 1.3001 & 0.3910 & 2.1322 & 0.6420 \\
& DepthPrompting~\cite{park2024depth}
& 2.9840 & 1.1430 & 2.3988 & 1.1290
& 1.8249 & 0.6240 & 2.9468 & 0.9869 \\
& OASIS-DC~\cite{cho2026oasis}
& \underline{1.4190} & \underline{0.5073}
& \underline{1.2830} & \underline{0.4001}
& 1.2455 & 0.3548
& \underline{1.5782} & 0.5540 \\
& UniDC~\cite{park2024simple}
& 1.6840 & 0.5217
& 1.3850 & 0.4073
& \underline{1.2238} & \underline{0.3386}
& 1.8378 & \underline{0.5406} \\
& LIFT (Ours)
& \textbf{1.3706} & \textbf{0.3750}
& \textbf{1.2767} & \textbf{0.3415}
& \textbf{1.1650} & \textbf{0.2992}
& \textbf{1.2655} & \textbf{0.3473} \\

\bottomrule
\end{tabular*}
\end{table*}

\subsection{Datasets and Evaluation Metrics}
We evaluate on NYUv2~\cite{silberman2012indoor} and KITTI-DC~\cite{uhrig2017sparsity}, representing indoor and outdoor depth completion, respectively. Following UniDC~\cite{park2024simple} and OASIS-DC~\cite{cho2026oasis}, we adopt the few-shot settings of 1, 10, and 100 annotated training frames, together with a 1-sequence setting. All training subsets are sampled from the official training split. For the 1-, 10-, and 100-shot settings, five subsets are independently sampled with different random seeds, and the reported results are averaged across the five runs.
For the 1-sequence setting, we follow UniDC~\cite{park2024simple} and report the average performance of models trained on each of the ten training sequences.

For NYUv2, images are resized to $240\times320$ and center-cropped to $228\times304$. Sparse inputs are generated by uniformly sampling 500 valid depth points per image, with random resampling during training and fixed sampling during evaluation. Performance is evaluated on the standard 654 test images over depths from 0.1 to 10~m. For KITTI-DC, inputs are bottom-center cropped to $352\times1216$, while all valid projected LiDAR measurements are retained as sparse inputs. Evaluation is conducted on the official 1,000-image cropped validation set over depths from 0.1 to 80~m.

We report Root Mean Squared Error (RMSE) and Mean Absolute Error (MAE), both measured in meters. Each metric is computed over valid ground-truth pixels for each image and averaged across the evaluation set.

\subsection{Implementation Details}
Our framework is implemented in PyTorch and trained on a single NVIDIA GeForce RTX 3090 GPU. We adopt the frozen Depth Anything V2 ViT-B/14 model~\cite{yang2024depth2} as the monocular depth foundation model to provide relative depth predictions and intermediate features. In the grid-based local alignment module, we use a grid stride of four pixels and select the $k$ nearest sparse measurements, with $k=10$ for NYUv2 and $k=20$ for KITTI-DC. For multi-scale depth propagation, we use features from layers 12, 9, 6, and 3 of the frozen MDE encoder to guide four successive coarse-to-fine stages, respectively. The projected features are bilinearly resized to the corresponding propagation resolutions: \((H_t,W_t)\), \((\lfloor H/4\rfloor,\lfloor W/4\rfloor)\), \((\lfloor H/2\rfloor,\lfloor W/2\rfloor)\), and \((H,W)\), where \(H_t\times W_t\) denotes the native encoder token-grid resolution and \(H\times W\) denotes the target depth resolution. For the training objective, we set $\lambda_{\mathrm{g}}$ to $0.1$ and $\lambda_{\mathrm{r}}$ to 5.0, 2.0, 0.2, and 0.2 for the 1-, 10-, 100-shot, and 1-sequence settings, respectively. The trainable modules are optimized using AdamW with a learning rate of $10^{-4}$, a weight decay of $10^{-3}$, and a batch size of 1. For 1-, 10-, and 100-shot training, we use 40, 500, and 1,500 iterations on NYUv2, and 40, 500, and 2,000 on KITTI-DC, respectively.

LIFT has 0.763M trainable parameters and runs in 0.144 s per image on KITTI-DC using an RTX 3090 GPU. The frozen MDE model, GLA, and propagation with residual refinement take 0.069, 0.036, and 0.039 s, respectively.

\begin{figure*}[!t]
\centering
\includegraphics[width=1.0\linewidth]{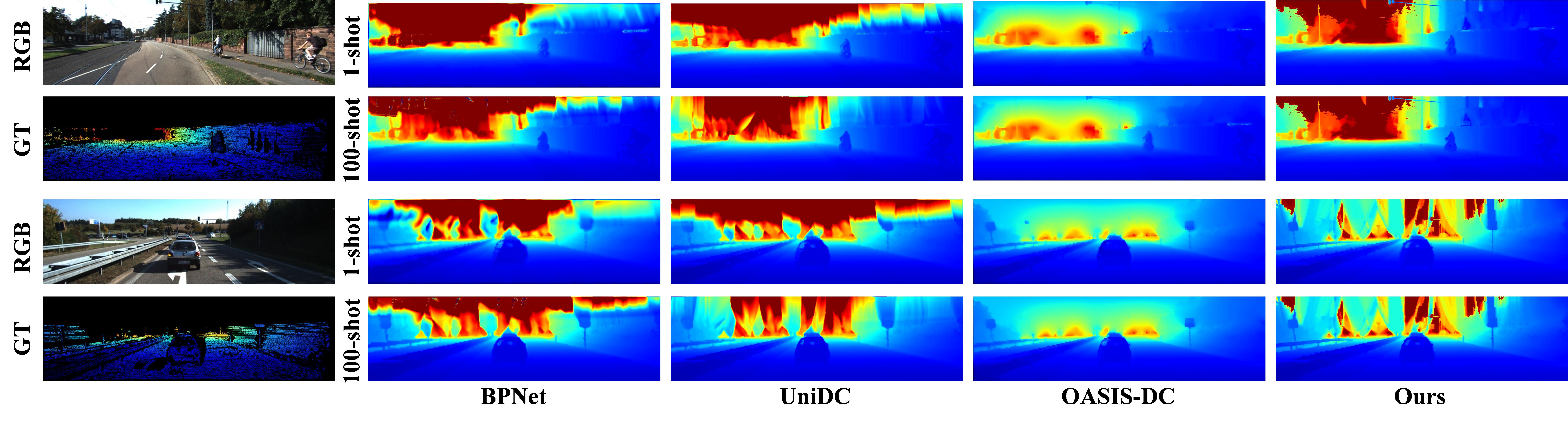}
\caption{Qualitative comparison on KITTI-DC under 1-shot and 100-shot supervision. The first column shows the RGB image and ground-truth depth, while the remaining columns show predictions from BPNet, UniDC, OASIS-DC, and our method. The top and bottom rows of predictions correspond to 1-shot and 100-shot training, respectively.
}
\label{fig:kitti}
\end{figure*}

\subsection{Comparison to State-of-the-Art Methods}
We compare LIFT with representative depth completion methods, including conventional propagation-based methods~\cite{cheng2018depth,park2020non,lin2023dyspn,zhang2023completionformer,tang2024bilateral}, recent prompting-based methods~\cite{park2024depth}, and state-of-the-art few-shot approaches~\cite{park2024simple,cho2026oasis}. 
As shown in Table~\ref{tab:fewshot_depth_completion}, LIFT consistently achieves the best results on both indoor and outdoor datasets across all supervision settings. In particular, LIFT attains RMSE/MAE of $0.1449/0.0619$ on NYUv2 and $1.3706/0.3750$ on KITTI-DC in the challenging 1-shot setting, substantially outperforming previous state-of-the-art methods. The consistent improvements across 1-, 10-, and 100-shot settings demonstrate the ability of the proposed framework to leverage pretrained depth priors under limited supervision. LIFT also achieves the best results under 1-sequence training, further demonstrating its robustness to both limited annotations and restricted scene diversity.

\begin{figure*}[!t]
\centering
\includegraphics[width=0.95\linewidth]{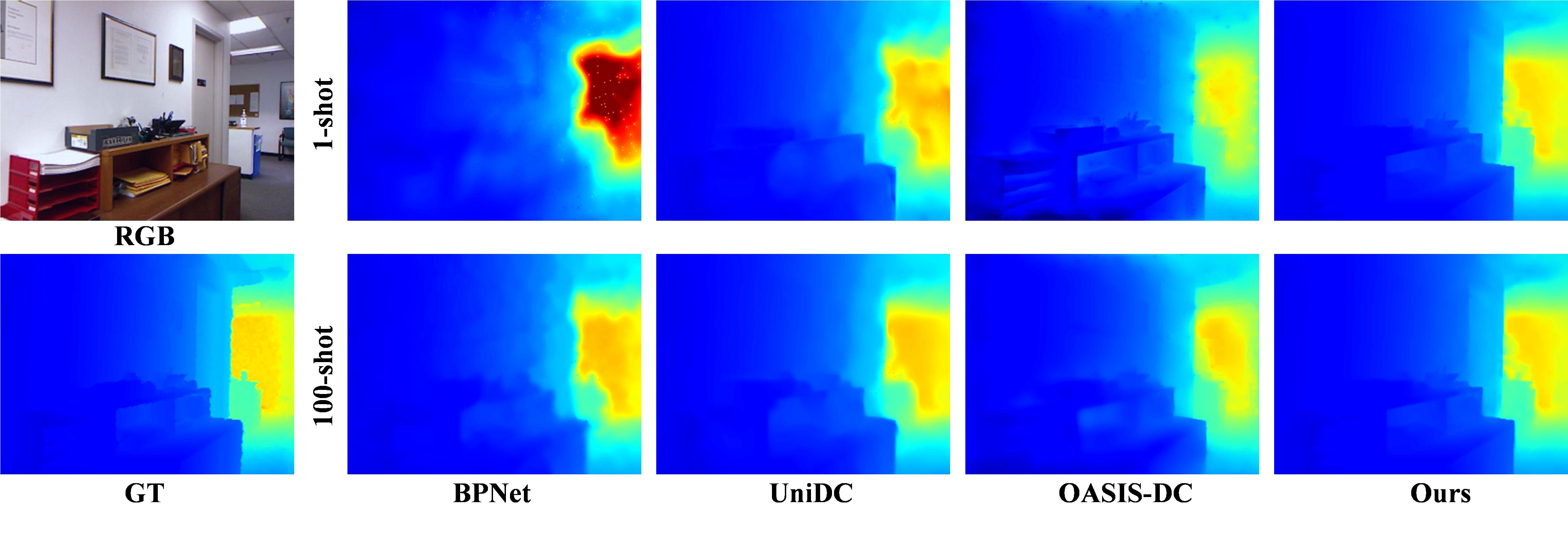}
\caption{Qualitative comparison on NYUv2 under 1-shot and 100-shot supervision. The first column shows the RGB image and ground-truth depth; the remaining columns compare BPNet, UniDC, OASIS-DC, and our method. The top and bottom prediction rows correspond to 1-shot and 100-shot training.}
\label{fig:nyuv2}
\end{figure*}

Fig.~\ref{fig:kitti} and Fig.~\ref{fig:nyuv2} provide qualitative examples under 1-shot and 100-shot supervision. On KITTI-DC, the predictions preserve clear depth transitions between the road and surrounding structures, while maintaining complete object shapes for objects such as riders and cars and recovering thin structures such as street lamps. On NYUv2, the predictions preserve the layout of furniture and capture depth transitions around doors and walls. The consistent overall geometry across the two supervision levels demonstrates the strong initialization provided by metric alignment, while the subsequent refinement modules further improve local structures and fine-grained details, even with extremely limited annotations.

\subsection{Ablation Studies}
\begin{table}[!tb]
    \centering
    \caption{Ablation study of the proposed components under 100-shot supervision on KITTI-DC. Global LS and Pixel LS denote global and pixel-wise least-squares fitting, respectively.}
    \label{tab:ablation}
    \small
    \setlength{\tabcolsep}{3pt}
    \renewcommand{\arraystretch}{1.12}
    \begin{tabular*}{\columnwidth}{@{\extracolsep{\fill}}lcccc@{}}
        \toprule
        Alignment & Propagation & Decoder & RMSE$\downarrow$ & MAE$\downarrow$ \\
        \midrule
        Global LS &   &  & 3.7918 & 1.7075 \\
        Global LS & $\ensuremath{\checkmark}$ & $\ensuremath{\checkmark}$ & 2.2355 & 0.6183 \\ 
				\midrule
				Pixel LS & & & 1.8138 & 0.3753 \\
				Pixel LS &$\ensuremath{\checkmark}$ & $\ensuremath{\checkmark}$ & 1.5530 & 0.3314\\
				\midrule
        GLA &   &   & 1.4293 & 0.4119 \\
        GLA & $\ensuremath{\checkmark}$ &  & 1.4111 & 0.3821 \\
        GLA &   & $\ensuremath{\checkmark}$ & 1.3481 & 0.3861 \\
        GLA & $\ensuremath{\checkmark}$ & $\ensuremath{\checkmark}$
            & \textbf{1.1650} & \textbf{0.2992} \\
        \bottomrule
    \end{tabular*}
\end{table}

\subsubsection{Effectiveness of Each Component}
We first conduct a series of ablation studies on KITTI-DC under the 100-shot setting to evaluate the contribution of each component. Table~\ref{tab:ablation} presents the framework-level ablation results. Starting from grid-based local alignment (GLA), adding multi-scale depth propagation provides a modest improvement by propagating reliable metric information to spatially and semantically consistent regions. In contrast, the residual decoder brings a larger gain by correcting local errors and recovering fine-grained details. Combining both modules further improves the performance to an RMSE of 1.1650 m and an MAE of 0.2992 m. This substantial gain suggests that propagation and residual refinement are complementary: propagation enhances the structural consistency of the depth estimate, while the residual decoder focuses on correcting remaining local errors and recovering fine-grained details.

\begin{table}[htbp]
\centering
\caption{Accuracy and runtime of different metric initialization methods on KITTI-DC and NYUv2. Global LS and Pixel LS denote global and pixel-wise least-squares fitting, respectively. Poisson denotes the Poisson-based reconstruction used in OASIS-DC~\cite{cho2026oasis} and LDCM~\cite{yu2026large}.}
\label{tab:metric_alignment}
\footnotesize
\setlength{\tabcolsep}{1.0pt}
\renewcommand{\arraystretch}{1.05}

\begin{tabular*}{\columnwidth}{@{\extracolsep{\fill}}lcccccc@{}}
\toprule
\multirow{2}{*}{\textbf{Alignment Methods}}
& \multicolumn{3}{c}{KITTI-DC}
& \multicolumn{3}{c}{NYUv2} \\
\cmidrule(lr){2-4} \cmidrule(lr){5-7}
& RMSE $\downarrow$
& MAE $\downarrow$
& T(s) $\downarrow$
& RMSE $\downarrow$
& MAE $\downarrow$
& T(s) $\downarrow$ \\
\midrule

DA-V2
& 65.3760 & 64.3420 & --
& 7.3239 & 7.2311 & -- \\
\midrule
Global LS
& 3.7918 & 1.7075 & 0.005
& 0.3362 & 0.2148 & 0.002 \\

Poisson \textcolor{gray!70}{\tiny OASIS-DC~\cite{cho2026oasis}}
& 1.5918 & 0.4184 & 0.199
& 0.2170 & 0.1269 & 0.062 \\

Poisson \textcolor{gray!70}{\tiny LDCM~\cite{yu2026large}}
& 1.5735 & 0.4765 & 0.290
& 0.2067 & 0.1098 & 0.199 \\

Pixel LS \textcolor{gray!70}{\tiny PriorDA~\cite{wang2026depth}}
& 1.8138 & \textbf{0.3753} & 0.254
& 0.1839 & \textbf{0.0607} & 0.029 \\

GLA (Ours)
& \textbf{1.4293} & 0.4119 & \textbf{0.036}
& \textbf{0.1552} & 0.0661 & \textbf{0.004} \\

\bottomrule
\end{tabular*}
\end{table}

\begin{figure}[!tb]
\centering
\includegraphics[width=1.0\linewidth]{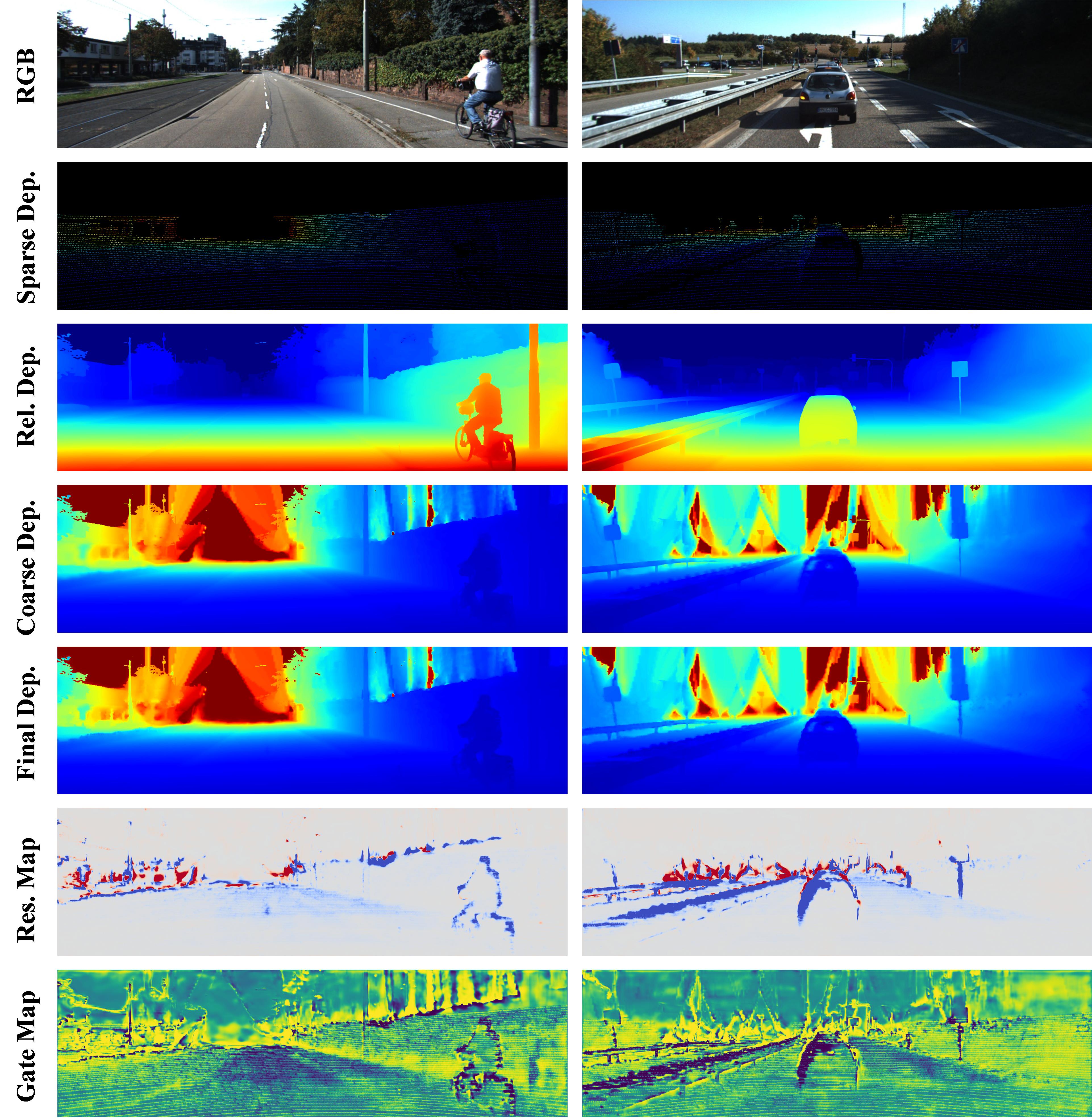}
\caption{Visualization of the outputs throughout the proposed pipeline. From top to bottom: RGB images, sparse depth, relative depth estimated by the MDE model, coarse metric depth aligned by GLA, final depth predictions, and the residual and fusion gate maps predicted by the residual decoder.}
\label{fig5}
\end{figure}

Fig.~\ref{fig5} illustrates the coarse aligned metric map, the predicted residual and fusion gate maps, and the final refined depth map, providing a qualitative view of the progressive refinement process. The GLA output already provides a reasonable metric depth initialization with consistent global geometry. The residual map is primarily concentrated around object boundaries and other local structural regions, indicating that the residual decoder focuses its refinement on areas where the initial prediction requires local corrections. By incorporating multi-scale depth propagation and residual refinement, the final prediction further improves structural coherence and recovers fine-grained details. These qualitative observations are consistent with the quantitative improvements reported in Table~\ref{tab:ablation}.
 
We additionally examine whether the proposed refinement modules are effective with different metric depth initializations. Applying both propagation and residual refinement consistently improves the results of global and pixel-wise alignment, reducing the RMSE from 3.7918 to 2.2355 and from 1.8138 to 1.5530, respectively. These results demonstrate that the proposed refinement modules are not tied to a specific initialization and can effectively enhance different metric depth estimates.

\subsubsection{Effectiveness of Grid-based Local Alignment}
Table~\ref{tab:metric_alignment} compares GLA with representative relative-to-metric depth alignment methods on KITTI-DC and NYUv2. Directly using the Depth Anything V2 predictions leads to large metric errors, while global least-squares fitting substantially reduces the errors but remains limited in modeling spatially varying scale and shift. Poisson-based reconstruction methods, including those used in OASIS-DC~\cite{cho2026oasis} and LDCM~\cite{yu2026large}, further improve the metric initialization by enforcing spatial consistency, while pixel-wise least-squares fitting~\cite{wang2026depth} provides more localized adjustment. Among these methods, GLA achieves the lowest RMSE on both KITTI-DC and NYUv2, with RMSEs of 1.4293 and 0.1552~m, respectively. Although pixel-wise least-squares fitting yields lower MAE, its RMSE is consistently higher than that of GLA, indicating that GLA is more effective at suppressing larger depth errors. Moreover, GLA requires only 0.036~s on KITTI-DC and 0.004~s on NYUv2, substantially less than the other local alignment methods. These results demonstrate that GLA provides an effective and efficient approach for converting relative depth predictions into reliable metric depth.

\begin{table}[!tb]
\centering
\caption{Propagation design ablations under 10-shot and 100-shot supervision on KITTI-DC.}
\label{tab:propagation}
\footnotesize
\setlength{\tabcolsep}{2pt}
\renewcommand{\arraystretch}{1.12}
\begin{tabular*}{\columnwidth}{@{\extracolsep{\fill}}lcccc@{}}
\toprule
\multirow{2}{*}{Settings} 
& \multicolumn{2}{c}{10-shot} 
& \multicolumn{2}{c}{100-shot} \\
\cmidrule(lr){2-3} \cmidrule(lr){4-5}
& RMSE $\downarrow$ 
& MAE $\downarrow$ 
& RMSE $\downarrow$ 
& MAE $\downarrow$ \\
\midrule

Full model    
& \textbf{1.2767} & 0.3415
& \textbf{1.1650} & \textbf{0.2992} \\

Residual propagation
& 1.3003 & \textbf{0.3408}
& 1.2004 & 0.3044 \\

RGB propagation
& 1.2945 & 0.3427
& 1.1804 & 0.3023 \\

Single scale
& 1.2963 & 0.3565
& 1.1770 & 0.3022 \\


\bottomrule
\end{tabular*}
\end{table}

\begin{table}[!tb]
\centering
\caption{Ablation of decoder conditioning inputs under 10-shot and 100-shot supervision on KITTI-DC.}
\label{tab:decoder}
\footnotesize
\setlength{\tabcolsep}{2pt}
\renewcommand{\arraystretch}{1.12}
\begin{tabular*}{\columnwidth}{@{\extracolsep{\fill}}ccccccc@{}}
\toprule
\multicolumn{3}{c}{Decoder Conditioning}
& \multicolumn{2}{c}{10-shot}
& \multicolumn{2}{c}{100-shot} \\
\cmidrule(lr){1-3}
\cmidrule(lr){4-5}
\cmidrule(lr){6-7}

$\{Z^{(l)}\}$
& $D_{\mathrm{out}}^{(L)}-\widehat{D}_0$
& $M$
& RMSE $\downarrow$
& MAE $\downarrow$
& RMSE $\downarrow$
& MAE $\downarrow$ \\
\midrule

\ensuremath{\checkmark} & \ensuremath{\checkmark} & \ensuremath{\checkmark}
& \textbf{1.2767} & \textbf{0.3415}
& \textbf{1.1650} & \textbf{0.2992} \\

\ensuremath{\checkmark} & &
& 1.3613 & 0.3794
& 1.3086 & 0.3776 \\

\ensuremath{\checkmark} & & \ensuremath{\checkmark}
& 1.3288 & 0.3640
& 1.2489 & 0.3386 \\

\ensuremath{\checkmark} & \ensuremath{\checkmark} &
& 1.2831 & 0.3443
& 1.1809 & 0.3121 \\

\bottomrule
\end{tabular*}
\end{table}

\subsubsection{Effectiveness of Multi-scale Depth Propagation}
Table~\ref{tab:propagation} evaluates different propagation designs under 10-shot and 100-shot supervision. The full multi-scale depth propagation consistently provides strong performance across both supervision settings. Replacing the propagated depth with residual propagation leads to a noticeable degradation, particularly under 100-shot supervision, indicating that explicit depth propagation provides more effective geometric cues for subsequent refinement. 
Using RGB guidance instead of foundation-feature guidance within the same multi-scale propagation structure increases RMSE and MAE in both settings, supporting the use of pretrained features for guiding propagation. Similarly, the single-scale variant performs worse on both metrics, supporting propagation across multiple spatial resolutions. 

\subsubsection{Effectiveness of Residual Decoder}
Table~\ref{tab:decoder} analyzes the inputs of the residual decoder under 10-shot and 100-shot supervision. The full decoder, conditioned on the multi-scale features $\{Z^{(l)}\}$, propagation correction $D_{\mathrm{out}}^{(L)}-\widehat{D}_0$, and sparse mask $M$, achieves the best performance. Using only foundation features or removing the propagated depth inputs degrades performance, supporting the benefit of geometric conditioning for residual refinement. The larger degradation under 100-shot supervision further suggests that these conditioning signals become more beneficial with increased supervision.

\section{Conclusions}
In this paper, we have presented LIFT, a few-shot depth completion framework that effectively leverages the priors of a pretrained monocular depth foundation model. LIFT first employs grid-based local alignment to convert relative depth predictions into spatially adaptive metric depth, and then progressively refines the aligned depth through multi-scale feature-guided propagation and a lightweight residual decoder. This design enables the framework to exploit both the global depth prior and local metric observations while requiring only a small number of annotated training samples. Extensive experiments on KITTI-DC and NYUv2 demonstrate the effectiveness of the proposed components and show that LIFT achieves competitive depth completion performance under extremely limited supervision. These results highlight the potential of foundation models as effective priors for few-shot metric depth completion.


\bibliographystyle{IEEEtran}
\bibliography{main}

\end{document}